\def\year{2020}\relax
\documentclass[letterpaper]{article} 
\usepackage{aaai20}  
\usepackage{times}  
\usepackage{helvet} 
\usepackage{courier}  
\usepackage[hyphens]{url}  
\usepackage{graphicx} 
\def\UrlFont{\rm}  
\usepackage{graphicx}  
\usepackage{multirow}
\usepackage{nicefrac}
\usepackage[export]{adjustbox}
\usepackage{amsmath,amssymb}
\DeclareMathOperator*{\argmin}{argmin}
\usepackage{caption}
\newcommand{\norm}[1]{\left\lVert#1\right\rVert}
\usepackage{enumitem}
\usepackage{tabularx}
\usepackage{booktabs}
\usepackage{makecell}
\usepackage{stfloats}
\usepackage[hang,flushmargin]{footmisc}

\newcolumntype{L}[1]{>{\raggedright\let\newline\\\arraybackslash\hspace{0pt}}m{#1}}
\newcolumntype{C}[1]{>{\centering\let\newline\\\arraybackslash\hspace{0pt}}m{#1}}
\newcolumntype{R}[1]{>{\raggedleft\let\newline\\\arraybackslash\hspace{0pt}}m{#1}}

\nocopyright

\makeatletter
\def\blfootnote{\xdef\@thefnmark{}\@footnotetext}
\makeatother

\title{Modeling Gestalt Visual Reasoning on the Raven's Progressive Matrices Intelligence Test Using Generative Image Inpainting Techniques}

\author{
     Tianyu Hua$^1$ and
     \Large \textbf{Maithilee Kunda} \\ 
     Electrical Engineering and Computer Science, Vanderbilt University, Nashville, TN, USA \\
     tianyu.hua@vanderbilt.edu, 
     mkunda@vanderbilt.edu
}
\begin{document}
\maketitle


\begin{abstract}
Psychologists recognize Raven's Progressive Matrices as a very effective test of general human intelligence.  While many computational models have been developed by the AI community to investigate different forms of top-down, deliberative reasoning on the test, there has been less research on bottom-up perceptual processes, like Gestalt image completion, that are also critical in human test performance.  In this work, we investigate how Gestalt visual reasoning on the Raven's test can be modeled using generative image inpainting techniques from computer vision.  We demonstrate that a self-supervised inpainting model trained only on photorealistic images of objects achieves a score of 27/36 on the Colored Progressive Matrices, which corresponds to average performance for nine-year-old children.  We also show that models trained on other datasets (faces, places, and textures) do not perform as well.  Our results illustrate how learning visual regularities in real-world images can translate into successful reasoning about artificial test stimuli.  On the flip side, our results also highlight the limitations of such transfer, which may explain why intelligence tests like the Raven's are often sensitive to people's individual sociocultural backgrounds.
\end{abstract}



\section{Introduction}
\begin{figure}[t]
    \centering
    \includegraphics[width=0.7\linewidth]{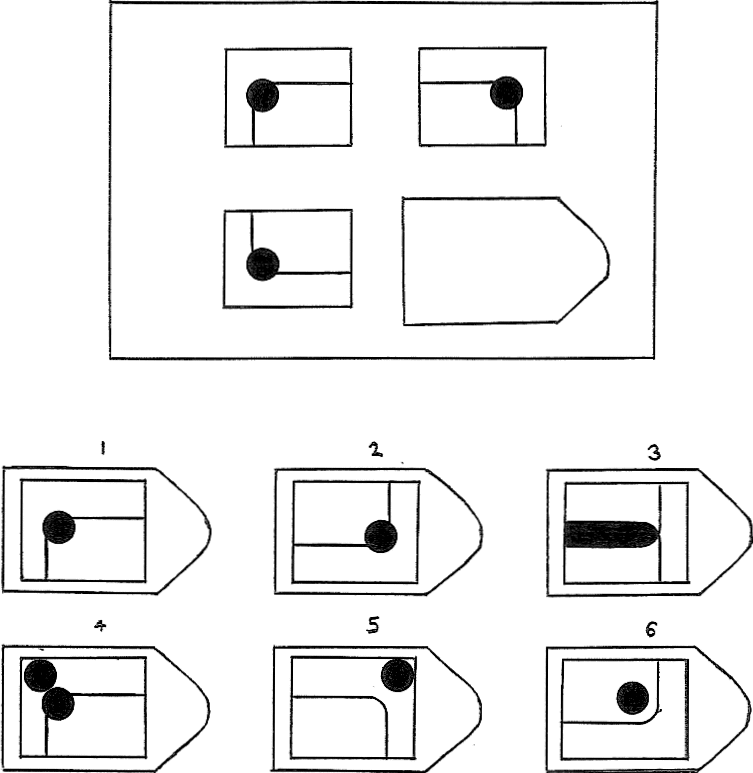}
    \caption{Example problem like those on the Raven's Progressive Matrices tests \cite{kunda2013computational}.}
    \label{fig:hand-made}
\end{figure}

Consider the matrix reasoning problem in Figure \ref{fig:hand-made}; the goal is to select the answer choice from the bottom that best fits in the blank portion on top.  Such problems are found on many different human intelligence tests \cite{roid1997leiter,wechsler2008wechsler}, including on the Raven's Progressive Matrices tests, which are considered to be the most effective single measure of general intelligence across all psychometric tests \cite{snow1984topography}. \blfootnote{$^1$Present affiliation: China University of Geosciences, Beijing.}

As you may have guessed, the solution to this problem is answer choice \#2.  While this problem may seem quite simple, what is interesting about it is that there are multiple ways to solve it.  For example, one might take a top-down, deliberative approach by first deciding that the top two elements are reflected across the horizontal axis, and then reflecting the bottom element to predict an answer--often called an Analytic approach \cite{lynn2004sex,prabhakaran1997neural}.  Alternatively, one might just ``see'' the answer emerge in the empty space, in a more bottom-up, automatic fashion--often called a Gestalt or figural approach.



While many computational models explore variations of the Analytic approach, less attention has been paid to the Gestalt approach, though both are critical in human intelligence.  In human cognition, Gestalt principles refer to a diverse set of capabilities for detecting and predicting perceptual regularities such as symmetry, closure, similarity, etc. \cite{wagemans2012century}.  Here, we investigate how Gestalt reasoning on the Raven's test can be modeled with generative image inpainting techniques from computer vision:

\begin{itemize}[nolistsep,noitemsep]
    \item We describe a concrete framework for solving Raven's problems through Gestalt visual reasoning, using a generic image inpainting model as a component.
    \item We demonstrate that our framework, using an inpainting model trained on photorealistic object images from ImageNet, achieves a score of 27/36 on the Raven's Colored Progressive Matrices test.
    \item We show that test performance is sensitive to the inpainting model's training data.  Models trained on faces, places, and textures get scores of 11, 17, and 18, respectively, and we offer some potential reasons for these differences.
\end{itemize}


\subsection{Background: Gestalt Reasoning}

\begin{figure}
    \centering
    \includegraphics[height=1.95cm]{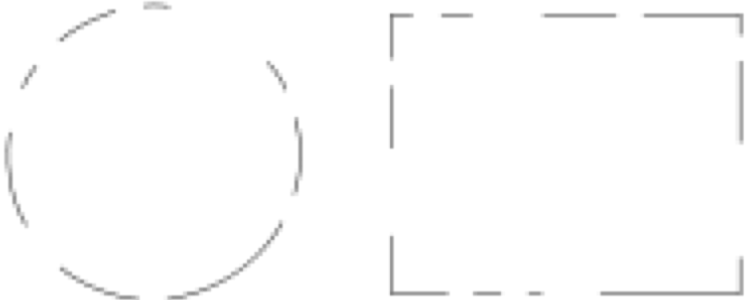} \hfill \includegraphics[height=1.95cm]{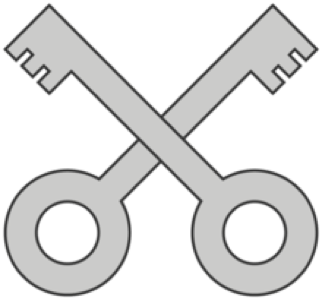}
    \caption{Images eliciting Gestalt ``completion'' phenomena.}
    \label{fig:gestalt}
\end{figure}

In humans, Gestalt phenomena have to do with how we integrate low-level perceptual elements into coherent, higher-level wholes \cite{wagemans2012century}.  For example, the left side of Figure \ref{fig:gestalt} contains only scattered line segments, but we inescapably see a circle and rectangle.  The right side of Figure \ref{fig:gestalt} contains one whole key and one broken key, but we see two whole keys with occlusion.

In psychology, studies of Gestalt phenomena have enumerated a list of principles (or laws, perceptual/reasoning processes, etc.) that cover the kinds of things that human perceptual systems do \cite{wertheimer1923untersuchungen,kanizsa1979organization}.  Likewise, work in image processing and computer vision has attempted to define these principles mathematically or computationally \cite{desolneux2007gestalt}.  

In more recent models, Gestalt principles are seen as emergent properties that reflect, rather than determine, perceptions of structure in an agent's visual environment.  For example, early approaches to image inpainting---i.e., reconstructing a missing/degraded part of an image---used rule-like principles to determine the structure of missing content, while later, machine-learning-based approaches attempt to learn structural regularities from data and apply them to new images \cite{schonlieb2015partial}.  This seems reasonable as a model of Gestalt phenomena in human cognition; after years of experience with the world around us, we see Figure \ref{fig:gestalt} (left) as partially occluded/degraded views of whole objects.


\subsection{Background: Image Inpainting}

Machine-learning-based inpainting techniques typically  either borrow information from within the occluded image itself \cite{bertalmio2000image,barnes2009patchmatch,ulyanov2018deep} or from a prior learned from other images \cite{hays2008scene,yu2018generative,zheng2019pluralistic}. The first type of approach often uses patch similarities to propagate low-level features, such as the texture of grass, from known background regions to unknown patches.  Of course, such approaches suffer on images with low self-similarity or when the missing part involves semantic-level cognition, e.g., a part of a face. 

The second approach aims to generalize regularities in visual content and structure across different images, and several impressive results have recently been achieved with the rise of deep-learning-based generative models.  For example, Li and colleagues (\citeyear{li2017generative}) use an encoder-decoder neural network structure, regulated by an adversarial loss function, to recover partly occluded face images. More recently, Yu and colleagues (\citeyear{yu2018generative}) designed an architecture that not only can synthesize missing image parts but also explicitly utilizes surrounding image feature as context to make inpainting more precise. In general, most recent neural-network-based image inpainting algorithms represent some combination of variational autoencoders (VAE) and generative adversarial networks (GAN) and typically contain an encoder, a decoder, and an adversarial discriminator. 



\subsubsection{Generative Adversarial Networks (GAN)}
Generative adversarial networks combine generative and discriminative models to learn very robust image priors \cite{goodfellow2014generative}. 
In a typical formulation, the generator is a transposed convolutional neural network while the discriminator is a regular convolutional neural network. During training, the generator is fed random noise and outputs a generated image. The generated image is sent alongside a real image to the discriminator, which outputs a score to evaluate how real or fake the inputs are. The error between the output score and ground truth score is back-propagated to adjust the weights. 

This training scheme forces the generator to produce images that will fool the discriminator into believing they are real images. In the end, training converges at an equilibrium where the generator cannot make the synthesized image more real, while the discriminator fails to tell whether an image is real or generated. Essentially, the training process of GANs forces the generated images to lay within the same distribution (in some latent space) as real images. 



\subsubsection{Variational autoencoders (VAE)}
Autoencoders are deep neural networks, with a narrow bottleneck layer in the middle, that can reconstruct high dimensional data from original inputs. The bottleneck will capture a compressed latent encoding that can then be used for tasks other than reconstruction. Variational autoencoders use a similar encoder-decoder structure but also encourage continuous sampling within the bottleneck layer so that the decoder, once trained, functions as a generator \cite{kingma2013auto}.



\subsubsection{VAE-GAN}

\begin{figure}[b]
    \centering
    \includegraphics[width=\linewidth]{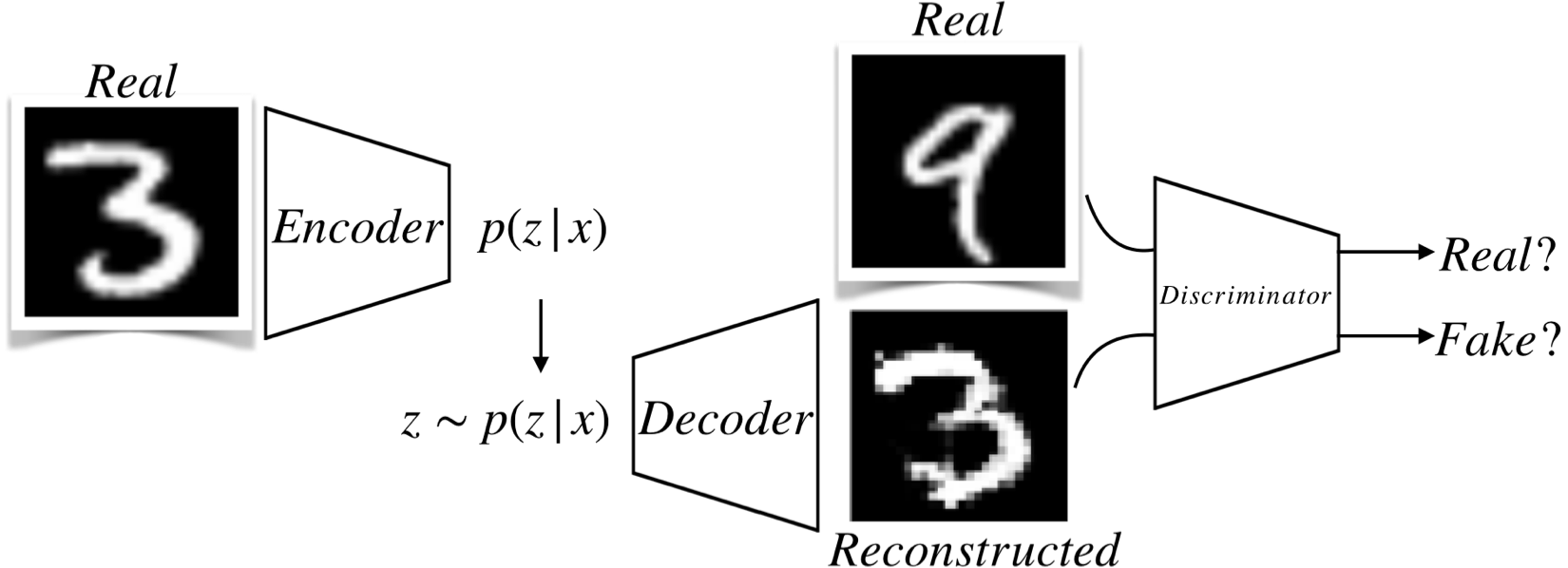}
    \caption{Architecture of VAE-GAN}
    \label{fig:vae-gan}
\end{figure}

While a GAN's generated image outputs are often sharp and clear, a major disadvantage is that the training process can be unstable and prone to problems 
\cite{goodfellow2014generative,Mao2016LeastSG}. Even if training problems can be solved, e.g., \cite{arjovsky2017wasserstein}, GANs still lack encoders that map real images to latent variables. Compared with GANs, VAE-generated images are often a bit blurrier, but the model structure in general is much more mathematically elegant and more easily trainable. To get the best of both worlds, Larsen and colleagues (\citeyear{larsen2015autoencoding}) proposed an architecture that attaches an adversarial loss to a variational autoencoder, as shown in Figure \ref{fig:vae-gan}. 


\begin{figure*}[b]
    \centering
    \includegraphics[width=0.8\linewidth,trim={0 1.15cm 0 1.25cm},clip]{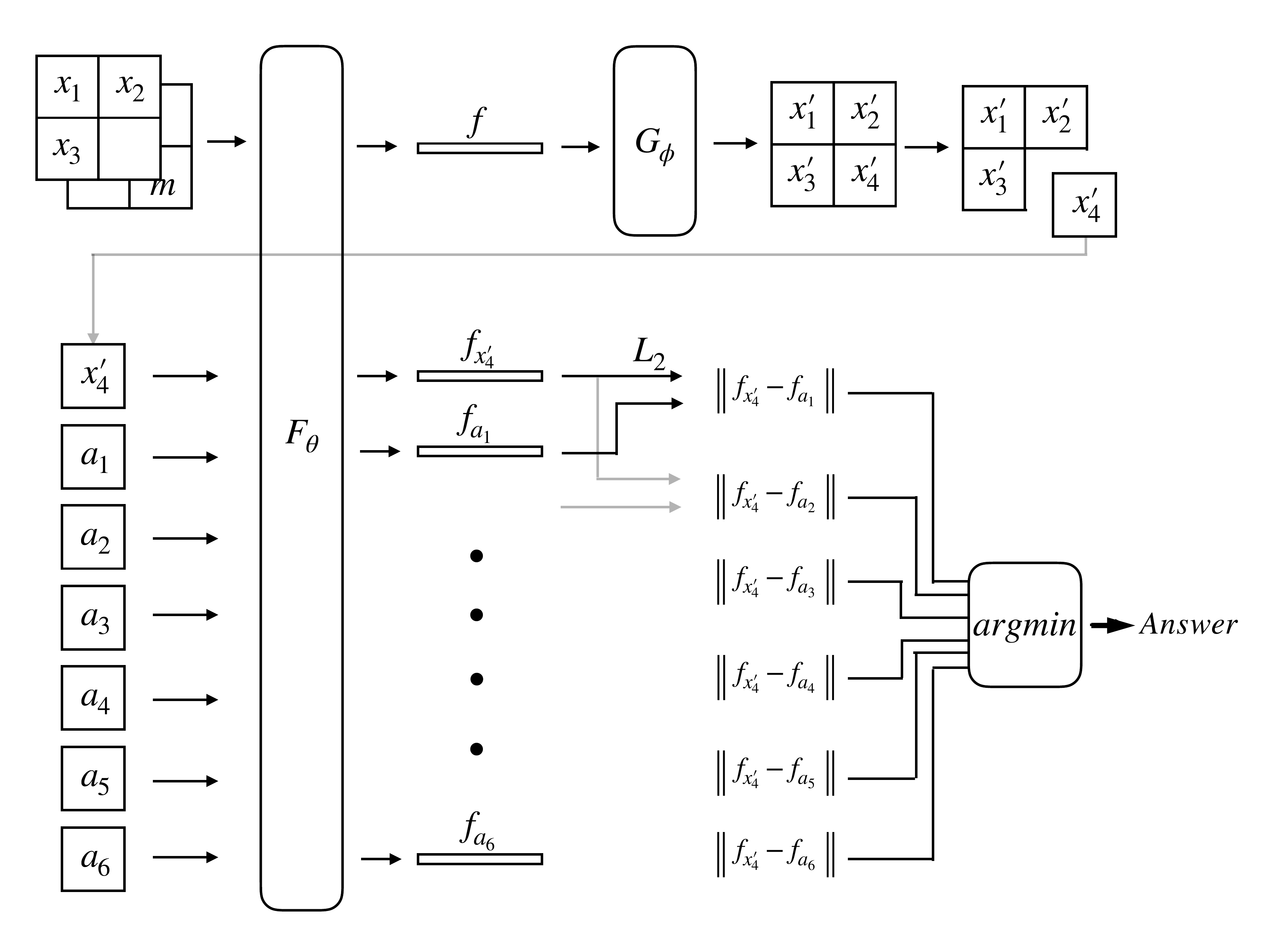}
    \caption{Reasoning framework for solving Raven's test problems using Gestalt image completion, using any pre-trained encoder-decoder-based image inpainting model.  Elements $x_1$, $x_2$, and $x_3$ from the problem matrix form the initial input, combined into a single image, along with a mask $m$ that indicates the missing portion. These are passed through the encoder $F_\theta$, and the resulting image features $f$ in latent variable space are passed into the decoder $G_\phi$.  This creates a new complete matrix image $X'$; the portion $x'_4$ corresponding to the masked location is the predicted answer to the problem.  This predicted answer $x'_4$, along with all of the answer choices $a_i$, are again passed through the encoder $F_\theta$ to obtain feature representations in latent space, and the answer choice most similar to $x'_4$ is selected as the final solution.}
    \label{fig:diagram}
\end{figure*}

\section{Our Gestalt Reasoning Framework}

\begin{figure*}
    \centering
    \includegraphics[width=\linewidth, keepaspectratio]{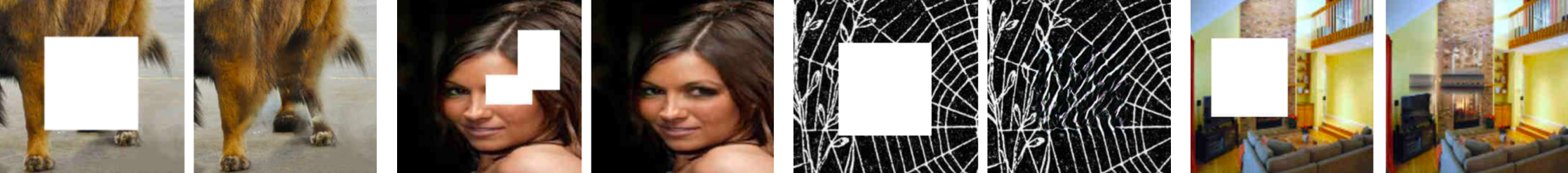}
    \caption{Examples of inpainting produced by same VAE-GAN model \cite{yu2018generative} trained on four different datasets.  Left to right: ImageNet (objects), CelebA (faces), Places (scenes), and DTD (textures).}
    \label{fig:data_example}
\end{figure*}

In this section, we present a general framework for modeling Gestalt visual reasoning on the Raven's test or similar types of problems.  Our framework is intended to be agnostic to any type of encoder-decoder-based inpainting model.  For our experiments, we adopt a recent VAE-GAN inpainting model \cite{yu2018generative}; as we use the identical architecture and training configuration, we refer readers to the original paper for more details about the inpainting model itself.  








Our framework makes use of a pre-trained encoder $F_\theta$ and corresponding decoder $G_\phi$ (where $\theta$ and $\phi$ indicate the encoder's and decoder's learned parameters, respectively). 
The partially visible image to be inpainted, in our case, is a Raven's problem matrix with the fourth cell missing, accompanied with a mask, which is passed as input into the encoder $F$. Then $F$ outputs an embedded feature representation $f$, which is sent as input to the generator $G$. Note that the learned feature representation $f$ could be of any form---a vector, matrix, tensor or any other encoding as long as it represents the latent features of input images.

The generator then outputs a generated image, and we cut out the generated part as the predicted answer. Finally, we choose the most similar candidate answer choice by computing the $L_2$ distance among feature representations of the various images (the prediction versus each answer choice), computed using the trained encoder $F$ again.


This process is illustrated in Figure \ref{fig:diagram}.  More concisely, let $x_1$, $x_2$, $x_3$, be the three elements of the original problem matrix, $m$ be the image mask, and $X$ be the input comprised of these four images. Then, the process of solving the problem to determine the chosen answer $y$ can be written as: 

$$
        y = \argmin_{k\in \mathbb{S}}\norm{F_\theta\biggl((G_\phi (F_\theta (X))_{ij})_{\substack{\frac{h}{2} < i \leq h \\ \frac{w}{2} < j \leq w}}\biggr) - F_\theta(a_k)}
$$

\noindent where h and w are height and width of the reconstructed image, and $\mathbb{S}$ is the answer choice space.

\subsection{Raven's Test Materials}

All Raven's problem images were taken from scans of official test booklets \cite{raven1998manual}.  We conducted experiments using two versions of the test: the Standard Progressive Matrices (SPM), intended for the general population, and the Colored Progressive Matrices (CPM), which is an easier test for children and lower-ability adults.  In fact, these two tests have substantial overlap: the CPM contains three sets labeled A, AB, and B, with 12 problems each, and the SPM contains five sets labeled A-E also with 12 problems each.  Sets A and B are shared across the two tests.  Problems increase in difficulty within and across sets.

Initial experiments showed that the inpainting models often failed to work when there was significant white space around the missing element, as in the problem in Figure \ref{fig:hand-made}.  Thus, when we fed in the matrix images as a combined single image, as in Figure \ref{fig:diagram}, we cropped out this white space.  This did change the appearance of problems somewhat, essentially squeezing together the elements in the matrix.

\subsection{Inpainting Models}

For our experiments, we used the same image inpainting model \cite{yu2018generative} trained on four different datasets.  The first model, which we call Model-Objects, we trained from scratch so that we could evaluate Raven's test performance at multiple checkpoints during training.  The latter three models, which we call Model-Faces, Model-Scenes, and Model-Textures, we obtained as pre-trained models \cite{yu2018generative}.  Details about each dataset are given below.

Note: The reader may wonder why we did not train an inpainting model on Raven's-like images, i.e., black and white illustrations of 2D shapes.  Our rationale follows the spirit of human intelligence testing: people are not meant to practice taking Raven's-like problems.  If they do, the test is no longer a valid measure of their intelligence \cite{hayes2015we}.  Here, our goal was to explore how ``test-naive'' Gestalt image completion processes would fare.  (There are many more nuances to these ideas, of course, which we discuss further in Related Work.)

\textbf{Model-Objects.}  The first model, Model-Objects, was trained on the Imagenet dataset \cite{russakovsky2015imagenet}.  We trained this model from scratch.  We began with the full ImageNet dataset containing $\sim$14M images non-uniformly spanning  20,000 categories such as ``windows,'' ``balloons,'' and ``giraffes.  The model converged prior to one full training epoch on the randomized dataset; we halted training around 300,000 iterations, with a batch size of 36 images per iteration.  The best Raven's performance was found at around 80,000 iterations, which means that the final model we used saw only about $\sim$3M images in total during training.

\textbf{Model-Faces.}  Our second model, Model-Faces, was trained on the Large-scale CelebFaces Attributes (CelebA) dataset \cite{liu2015deep}, which contains around 200,000 images of celebrity faces, covering around 10,000 individuals.

\textbf{Model-Scenes.} Our third model, Model-Scenes, was trained on the Places dataset \cite{zhou2017places}, which contains around 10M images spanning 434 categories, grouped into three macro-categories: indoor, nature, and urban.

\textbf{Model-Textures.} Our fourth model, Model-Textures, was trained on the Describable Textures Dataset (DTD) \cite{cimpoi2014describing}, which contains 5640 images, divided into 47 categories, of textures taken from real objects, such as knitting patterns, spiderwebs, or an animal's skin. 


\begin{figure}[h]
    \centering
    \includegraphics[width=0.9\linewidth,trim={0.3cm 0 0 0},clip]{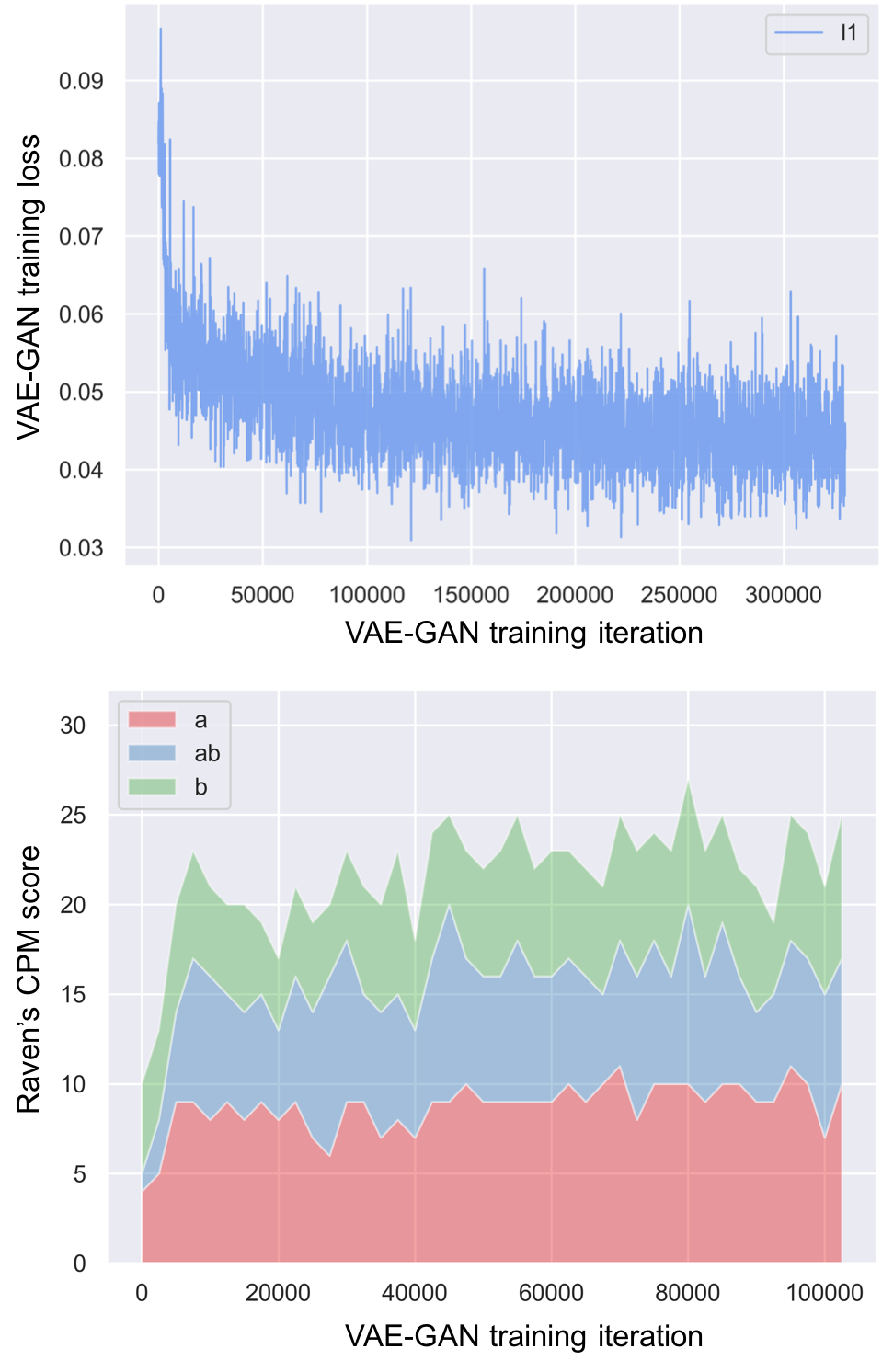}
    \caption{Image inpainting loss (top) and CPM performance (bottom) during training of Model-Objects.}
    \label{fig:trainingcurves}
\end{figure}

\section{Results}

Figure \ref{fig:trainingcurves} shows results over training time for Model-Objects.  The top plot shows the loss function that is being trained for image inpainting; the model seems to settle into a minimum around 200,000 iterations.  
The bottom plot shows CPM performance as a function of training, divided into sets A, AB, and B.  The model relatively quickly rises above chance performance, which would be an expected score of 6 in total (from 36 problems each having 6 answer choices).  

In fact, we noticed that the randomly initialized model actually appears to do a bit better than chance; after numerous runs, the average starting score was around 8/36.  We believe this can be attributed to intrinsic structure-capturing abilities of the convolutional neural network structure \cite{ulyanov2018deep}.

After $\sim$80,000 iterations, CPM performance does not change other than local variations.  
For the rest of our analyses, we used the model snapshot at the point when it reached peak performance of 27/36 correct.  While this yields an optimistic estimate of performance, we chose this approach in keeping with our goal of investigating what sort of Gestalt transfer would even be possible using a model that had never seen Raven's problems before.

Now we compare results across the four models: Model-Objects trained as above, and pre-trained versions of Model-Faces, Model-Scenes, and Model-Textures.

\begin{figure}
	\centering
	\includegraphics[width=\linewidth,trim={0.65cm 1.25cm 0.5cm 0.25cm},clip]{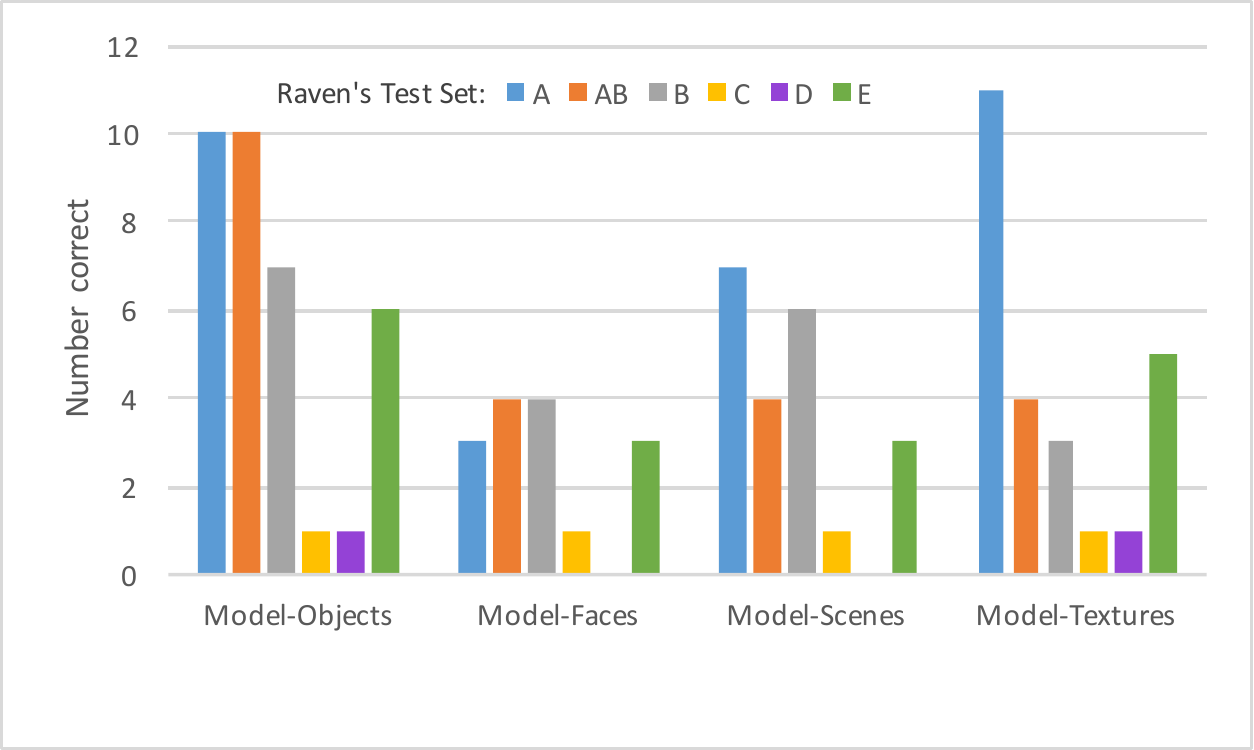}
	\caption{Results for each model on each set of the Raven's CPM (A, AB, and B) and SPM (A-E).}
	\label{fig:setresults}
\end{figure}

Figure \ref{fig:setresults} shows scores achieved by each of the four models on each of the six sets of Raven's problems.  As seen in this plot, Model-Objects performs better than any of the other models overall, though Model-Textures does a smidgeon better on Set A (which contains very texture-like problems, so this result makes sense).

None of the models do very well on sets C or D, performing essentially at chance (these problems have 8 answer choices, so chance $\sim$1.5 correct.  Interestingly, Model-Objects was the only one that consistently generated answers to all problems; the other three models often generated blank images to problems in sets C and D.  We are not sure why this occurred.  All of the models do rather surprisingly well on set E, which is supposed to be the hardest set of all.

\begin{figure}[h]
	\centering
	\includegraphics[width=\linewidth,trim={0.6cm 0.85cm 0.4cm 0.35cm},clip]{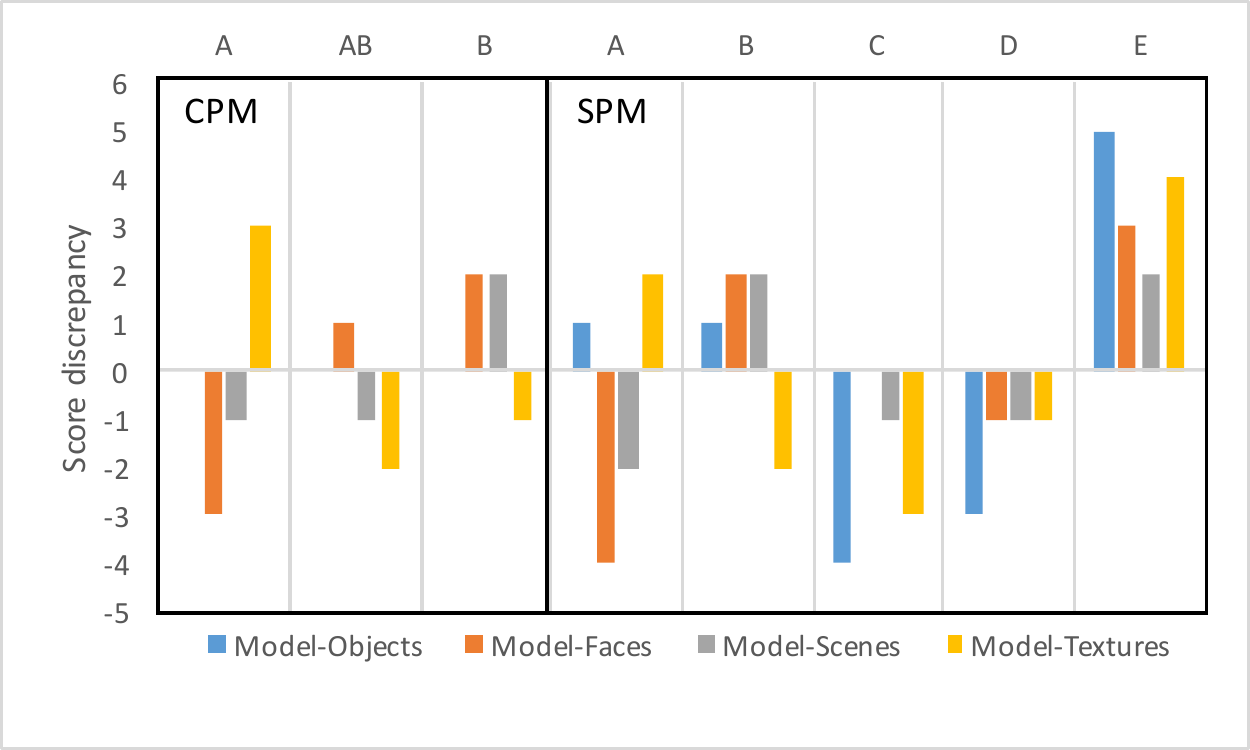}
	\caption{Per-set score discrepancies between each model and human norms for same total scores on CPM and SPM.}
	\label{fig:discreps}
\end{figure}

Figure \ref{fig:discreps} shows values called ``score discrepancies.''  When a person takes a Raven's test, the examiner is supposed to check the per-set composition of their score against normative data from other people who got the same total score.  So, for example, a score of 27 on the CPM has norms of 10, 10, and 7 for sets A, AB, and B, respectively, which is exactly what Model-Objects scored.  (This is why there are no blue bars appearing in the CPM portion of this plot.)  This means that Model-Objects was essentially subject to the same difficulty distribution as other people taking the test.  

In contrast, if we look at the SPM results, the models do worse than they should have on sets C and D, and better than they should have on set E.  This means that the difficulty distributions experiences by the models are not the same as what people typically experience.

\begin{figure*}
	\centering
	\includegraphics[width=\linewidth]{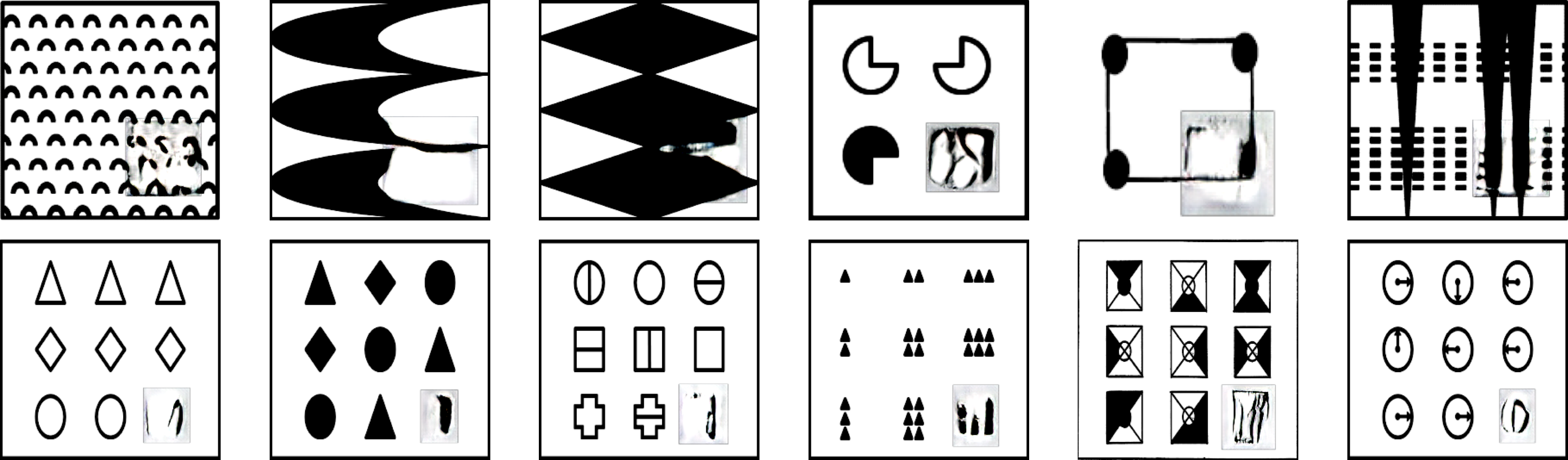}
	\caption{Images generated using Model-Objects for a variety of Raven's-like sample problems.}
	\label{fig:samples}
\end{figure*}

Figure \ref{fig:samples} shows examples of Model-Objects results from various sample problems.  (Actual Raven's problems are not shown, in order to protect test security.)  Some results are surprisingly good, given that the model was only trained on real-world color photographs.

Interestingly, when we inspected results from the Raven's test, the model generates what look like poor image guesses for certain problems, for example on some of the more difficult problems in set E, but then still chooses the correct answer choice.  This could be some form of lucky informed guessing, or, it could be that the image representations in latent space are actually capturing some salient features of the problem and solution.

\section{Discussion and Related Work}

Over the decades, there have been many exciting efforts in AI to computationally model various aspects of problem solving for Raven's matrix reasoning and similar geometric analogy problems, beginning with Evans' classic ANALOGY program \cite{evans1968program}.  In this section, we review some major themes that seem to have emerged across these efforts, situate our current work within this broader context, and point out important gaps that remain unfilled.



Note that our discussion does not focus heavily on absolute test scores.  Raven's is not now (and probably never will be) a task that is of practical utility for AI systems in the world to be solving well, and so treating it as a black-box benchmark is of limited value.  However, the test has been and continues to be enormously profitable as a research tool for generating insights into the organization of intelligence, both in humans and in artificial systems.  We feel that the more valuable scientific knowledge from computational studies of Raven's problem solving has come from systematic, within-model experiments, which is also our aim here.

\textbf{Knowledge-based versus data-driven.}  Early models took a knowledge-based approach, meaning that they contained explicit, structured representations of certain key elements of domain knowledge.  For example, Carpenter and colleagues (\citeyear{carpenter1990one}) built a system that matched relationships among problem elements according to one of five predefined rules. Knowledge-based models tend to focus on what an agent does with its knowledge during reasoning \cite{rasmussen2011neural,kunda2013computational,strannegaard2013anthropomorphic,lovett2017modeling}; where this knowledge might come from remains an open question.  

On the flip side, a recently emerging crop of data-driven models extract domain knowledge from a training set containing example problems that are similar to the test problems the model will eventually solve \cite{hoshen2017iq,barrett2018measuring,hill2019learning,steenbrugge2018improving,van2019disentangled,zhang2019raven}.  Data-driven models tend to focus on interactions between training data, learning architectures, and learning outcomes; how knowledge might be represented in a task-general manner and used flexibly during reasoning and decision-making remain open questions.

Our model of Gestalt visual reasoning falls into an interesting grey area between these two camps.  On the one hand, the model represents Gestalt principles implicitly, as image priors in some latent space, and these priors are learned in a data-driven fashion.  On the other hand, unlike all of the above data-driven models, our model does \textit{not} train on anything resembling Raven's problems.  In that sense, it is closer to a knowledge-based model in that we can investigate how knowledge learned in one setting (image inpainting) can be applied to reason about very different inputs.

\textbf{Constructive matching versus response elimination.}  Another interesting divide among Raven's models has to do with the overall problem-solving strategy.  A study of human problem solving on geometric analogy problems found that people generally use one of two strategies: they come up with a predicted answer first, and then compare it to the answer choices---constructive matching---or they mentally plug each answer choice into the matrix and choose the best one---response elimination \cite{bethell1984adaptive}.

Knowledge-based models have come in both varieties; all of the data-driven models follow the response-elimination approach.  Our model uses constructive matching, which we feel is an interesting capability given that the system is not doing any deliberative reasoning (per se) about what should go in the blank space.

\begin{figure}[h]
    \centering
    \includegraphics[width=\linewidth]{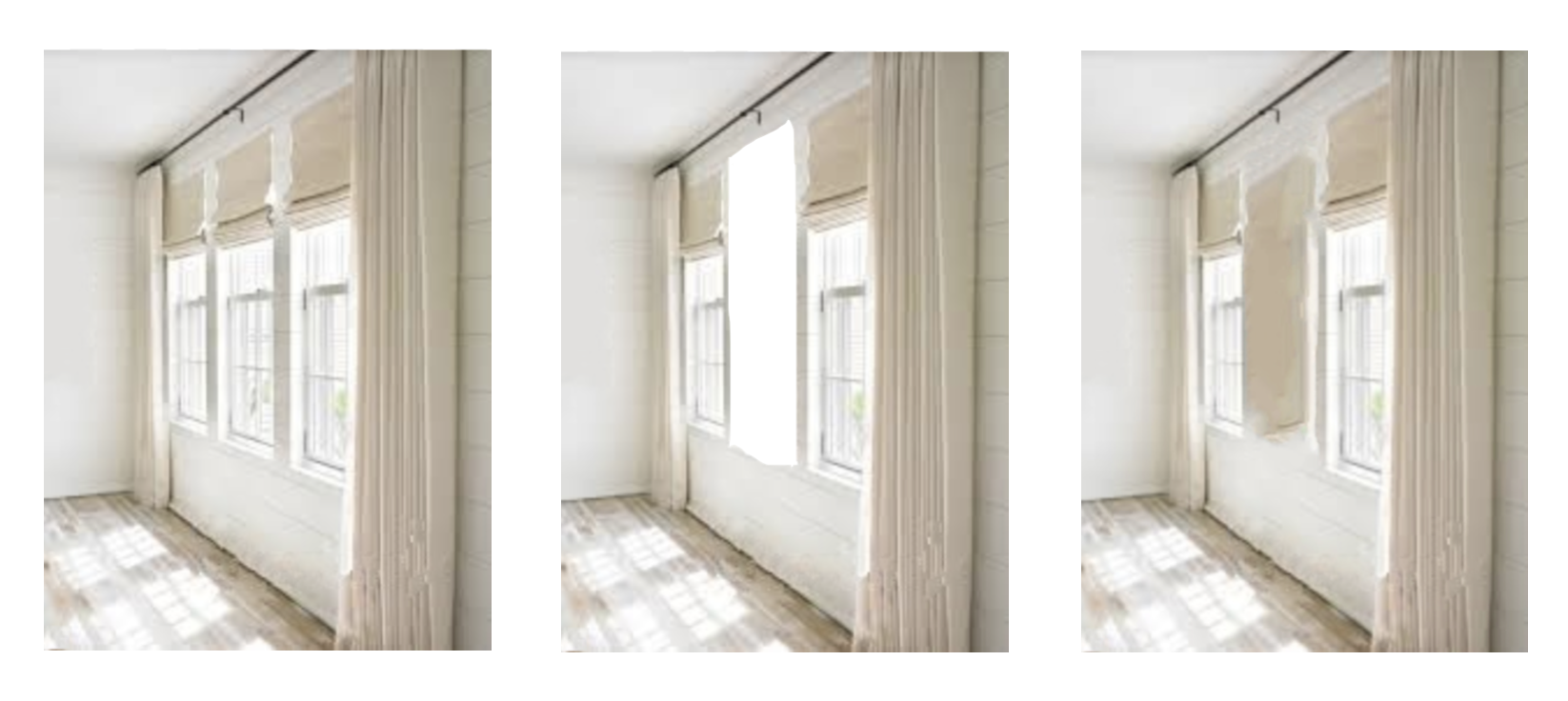}
    \caption{Model-Objects performing inpainting on a row of windows, with the original image on the left, the masked image in the center, and the inpainted image on the right.  Note the phantom reflection in the inpainted image.  This type of relational, commonsense reasoning requires going beyond a purely Gestalt approach.}
    \label{fig:arowofwindows}
\end{figure}

\textbf{Open issues.}  Our Gestalt model certainly has limitations, as illustrated in the results section.  (See Figure \ref{fig:arowofwindows} for another example.)  However, our investigations highlight a form of human reasoning that has not been explored in previous Raven's models.  How are Gestalt principles learned, and how do specific types of visual experiences contribute to a person's sensitivity to regularities like symmetry or closure?

One fascinating direction for future work will be to explore these relationships in more detail, and perhaps shed light on cultural factors in intelligence testing.  For example, would a model trained only on urban scenes (which contain lots of corners, perfect symmetries, and straight lines) do better on Raven's problems than a model trained only on nature scenes?

Finally, two major open issues for AI models of intelligence tests in general are: metacognitive strategy selection, and task learning.  Most AI models tend to adopt a single strategy and see how far its performance can be pushed.  However, for humans, a major part of the challenge of intelligence testing is figuring out what strategy to use when, and being able to adapt and switch strategies as needed.

In the context of our work, we aim to integrate our Gestalt approach with other, more deliberative reasoning approaches to begin to address this issue.  This will introduce many challenges related to having to determine confidence in an answer, planning and decision making, etc.

Relatedly, as with many tasks and systems in AI, previous work on Raven's and other intelligence tests has required the AI system designers to specify the task, its format, goal, etc. for the system.  Humans sit down and are given verbal or demonstration-based instructions, and must learn the task, how to represent it internally, and how and what procedures to try.  This kind of task learning \cite{laird2017interactive} remains a key challenge for AI research in intelligence testing.

\section*{Acknowledgments}

This work was funded in part by the National Science Foundation, award \#1730044.


\bibliography{references}
\bibliographystyle{aaai}

\end{document}